\documentclass[aps,pre,reprint,final,10pt,twocolumn,oneside,nobalancelastpage,
flushbottom,groupedaddress,superscriptaddress,showpacs,showkeys,
amsfonts,amssymb,amsmath,longbibliography]{revtex4-1}
\usepackage[T2A]{fontenc}
\usepackage[utf8]{inputenc}

\usepackage[left=2cm,right=1.5cm,top=1.5cm,bottom=2cm]{geometry}
\usepackage{color,xcolor}
\usepackage{graphicx}
\usepackage{epstopdf}
\usepackage{hyperref}
\usepackage{cleveref}
\usepackage{natbib}
\usepackage{subcaption}
\usepackage{multirow}
\usepackage{array}

\usepackage{bm}

\begin{document}
\title{ECG Segmentation by Neural Networks: Errors and Correction}

\thanks{The study was supported by the Ministry of Education and Science of Russia (Project No. 14.Y26.31.0022).}

\author{Iana Sereda}
\affiliation{Department of Control Theory, Nizhny Novgorod State University, Gagarin Av. 23, Nizhny Novgorod, 603950, Russia}
	
\author{Sergey Alekseev}
\affiliation{Department of Control Theory, Nizhny Novgorod State University, Gagarin Av. 23, Nizhny Novgorod, 603950, Russia}
	
\author{Aleksandra Koneva}
\affiliation{Department of Control Theory, Nizhny Novgorod State University, Gagarin Av. 23, Nizhny Novgorod, 603950, Russia}
	
\author{Roman Kataev}
\affiliation{Department of Control Theory, Nizhny Novgorod State University, Gagarin Av. 23, Nizhny Novgorod, 603950, Russia}
	
\author{Grigory Osipov}
\affiliation{Department of Control Theory, Nizhny Novgorod State University, Gagarin Av. 23, Nizhny Novgorod, 603950, Russia}

\begin{abstract}
	\begin{center}    {\bf Abstract}    \end{center}
{
 In this study we examined the question of how error correction occurs in an ensemble of deep convolutional networks, trained for an important applied problem: segmentation of Electrocardiograms(ECG). We also explore the possibility of using the information about ensemble errors to evaluate a quality of data representation, built by the network. This possibility arises from the effect of distillation of outliers, which was demonstarted for the ensemble, described in this paper.
}
		
\end{abstract}

\keywords
{convolutional neural networks, cardiac cycle, segmentation, ensemble, outliers, errors}
\maketitle

\section{Introduction}
Correction of errors of Artificial Intelligence (AI) systems is recognized as one of the main problems in the AI-based technical revolution \cite{gorban2018correction}. The effect of error correction often appears in ensembles of neural networks: it is known that, in most cases, an ensemble can improve the effectiveness of the base network \cite{huang2017snapshot}. The creation of an ensemble of models is widely used in modern machine learning as the last step of the working pipeline. However, it is difficult to predict which mistakes the ensemble can eliminate from the basic model and which can not.

This problem of possible mistakes of the trained model remains relevant because the representation of the data learned by the neural network is difficult to interpret \cite{zhang2018visual}. The reliability of a neural network is directly connected to the quality of the internal data representation that it has built. In the context of medical tasks, the problem of analyzing the quality of representation (and fixing the flows in it) is especially important due to the peculiarities of medical datasets: different pathologies are often represented by a small number of samples, while variants close to the norm may occur too often \cite{yap2014application}. However, it is the pathological cases that are most important. 
 
Imbalance of the data set often leads to the situation there formal quality metrics can give unreasonably good result, while the network could not cover all important aspects of data well. It is not always possible to combat data imbalances with well-known methods (such as, for example, oversampling), because is not always clearly visible which particular classes require balancing. We illustrate the above problem using the example of the ECG markup task: all meaningful components of cardiac cycle ( P-wave, T-wave and the QRS complex)  are roughly balanced in most of ECGs due to the periodicity of ECG structure. But the task itself contains imbalance, because the dataset is not balanced for diseases. Diseases change the morphology of the components of the cardiac cycle in different ways, so the representation built by the neural network must contain information of how the cardiac cycle looks like for every pathology presented in the dataset. When creating a quality metric for an arbitrary task, it is difficult to take into account the imbalance for all the hidden factors of influence existing in this task. In this paper, we use data on how exactly the ensemble corrects the errors of the underlying network in order to conclude about the quality of data representation received by the network.

One of the ways to investigate the reliability of the data representation received within the network is to use adversarial examples \cite{arnab2017robustness}. Another common approach to analyzing the quality of the representation of deep networks is based on the visualization of the learned attributes of different levels \cite{yosinski2015understanding}. For models with attention, attention visualization can be used \cite{xu2015show}. A new direction is to find a metric for evaluating the degree of disentanglement in representations \cite{eastwood2018framework}. Other methods can be found in a survey \cite{zhang2018visual}.
 
This paper is organized as follows. First, we describe the details of training of a convolutional network on an ECG segmentation dataset. Results are averaged over multiple runs and are measured by means of the usual metrics adopted to measure quality in this medical task (see section \ref{sec::eval}) . In the last sections we investigate the error correction by an iteratively formed network ensemble and demonstrate the effect of distilling such pathological cases to which the network of a given architecture is most vulnerable, given that data set. 

The project’s code is publiclly available at: https://github.com/Namenaro/ecg\_segmentation

\section{Dataset}
\begin{figure}[h!]
\centering
\includegraphics[scale=0.5]{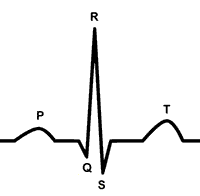}
\caption{Schematic cardiac cycle}
\label{pqrst}
\end{figure}

LUDB\cite{2018arXiv180903393K} is an open access dataset, containing ECG recordings of 200 unique patients. Each recording is represented by a 10-second signal registered from twelve leads with a sampling rate of 500 Hz. An expert's annotation is provided for each patient, annotating the three segments of the cardiac cycle: P, QRS and T. The proper detection of these waves/complexes is essential for ECG-based diagnostics of the cardiovascular system. A schematic representation of the cardiac cycle is shown on fig. \ref{pqrst}.
 A significant part of the dataset is represented by healthy cases, and the remaining part covers a wide range of different pathologies of the cardiovascular system (different heart rhythm types, conduction abnormalities, repolarization abnormalities and so on). The dataset also contains ECGs with varying degree of noise.
 
 \section{Data preparation}
\begin{figure}[h!]
\centering
    \begin{subfigure}[t]{0.5\textwidth}
    \includegraphics[scale=0.22]{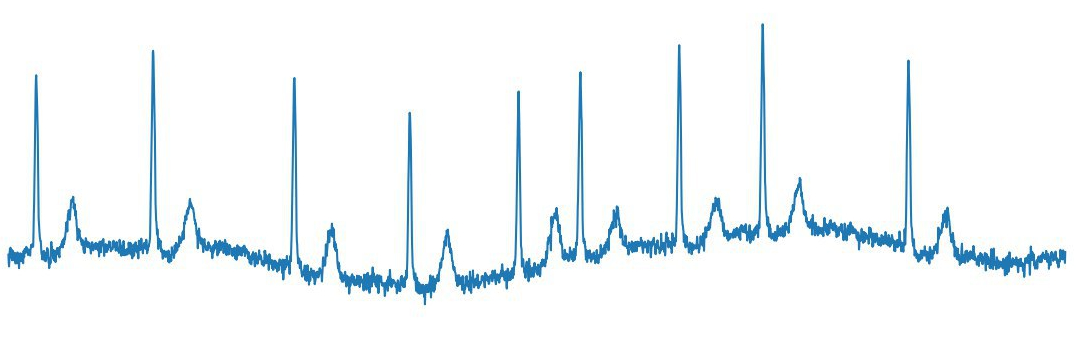}
   
    \label{before_drift_elim}
    \end{subfigure}
    \\ 
    \begin{subfigure}[t]{0.5\textwidth}
    \includegraphics[scale=0.22]{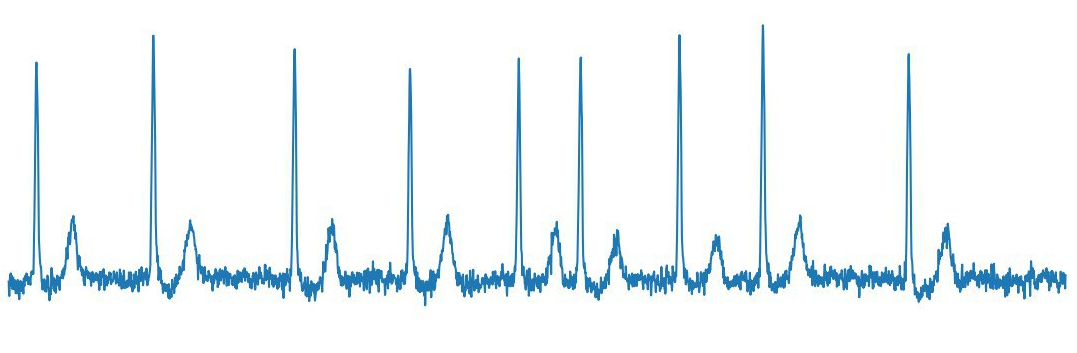}
     
    \label{after_drift_elim}
    \end{subfigure}
\caption{BW removal example. ECG signal before and after processing is shown at the top and bottom respectively}
\label{drift_example}
\end{figure}

ECG preprocessing has consisted of Baseline wander(BW) removal, which is a conventional first step in ECG processing for most applications\cite{de2004automatic}.
Baseline wander is a low-frequency ECG artifact, which may be caused by patients’ breath or movement \cite{lenis2017comparison} and holds no diagnostic information. ECG was filtered with two median filters as described in \cite{de2004automatic}. The resulting ECG is shown in fig. \ref{drift_example}.
High frequency noise was not removed, and no augmentation was performed.

\section{Base neural network}\label{sec::base}
\begin{figure}[h!]
\centering
\includegraphics[scale=0.59]{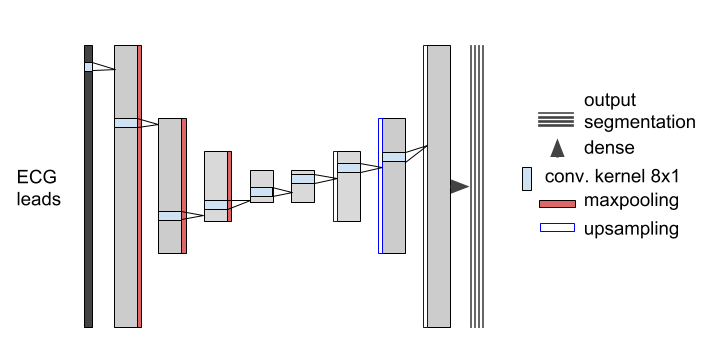}
\caption{An 8-layer convolutional architecture for which this error analysis was performed}
\label{convnet}
\end{figure}

The architecture of the base neural network to be used is shown in fig. \ref{convnet}.
It was shown that the convolutional neural network architecture is well-suited for many real world problems, such as medical signal processing and analysis, including ECG signal annotation\cite{rajpurkar2017cardiologist}.

The output signal is comprised of four channels: one contains the annotation mask for the P peak, the second for the QRS complex, the third channel has the mask for the T peak. The fourth channel is auxiliary, and is equivalent to background commonly found in the segmentation problem.
Examples of input and their corresponding output signals are shown in pictures \ref{delin_example}, \ref{worst}.
Softmax function is applied to each pixel of the annotation at the last layer of the network, which is fully connected. This function works with various channels and therefore lays an a priori assumption that activity on one of the output channels must tend to exclude any activity in the other channels.

The end result represents binary masks for all types of peaks. For each channel, the mask is generated individually based on the output signal (pic. \ref{delin_example}) as follows: for each point in time, a “winner” channel is selected (one which has the largest value at that time point), which gains 1, the rest of the channels gain 0.

\section{Training}
Training was conducted on one ECG lead, because experiments have shown that the addition of the remaining ECG leads does not bring any statistically significant improvements to the results of the neural network.
The data set was split into training (134 patients) and test (66 patients) parts. 
No post-processing was performed on the resulting segmentation.
During training, the mini-batches are formed out of randomly selected 6-second intervals. Other parameters of training are: the chosen optimizer is RMSProp and the loss function is categorical cross entropy. 

\section{Evaluation}\label{sec::eval}
In this work, we define annotation as designation of the beginning and end points for each of the waves/complexes specified. To evaluate the annotation quality for a particular type of points, (such as the P-wave starting points) we employ an algorithm that works as follows: for each point of this type on the doctor's annotation, the algorithm looks for the corresponding point on the network's annotation.

If a corresponding point is found in the specified neighborhood of the doctor's point, then we count the network’s decision as valid (True Positive, TP). In this case the error value is calculated as the distance between the point in the doctor's annotation and the corresponding point in the network's annotation. 

If a point specified by the network does not exist on the doctor's annotation in the specified neighbourhood, we count the answer as false positive (FP). Should the network be unable to locate a point which is present in the doctor’s annotation, then the answer is marked as false negative (FN).

The permitted neighbourhood is calculated adaptively depending on the patient's heart rate: for a heart rate of 70 BPM, the radius of the permitted neighbourhood was chosen to be 150 ms. Then, the size of this neighbourhood is decreased linearly based on the length of the cardiac cycle. An interval of 150 milliseconds was selected in accordance with ANSI/AAMI-EC57:1998.

The following quality metrics are commonly used for ECG annotation evaluation: \\
\begin{itemize}
\item $m$ -- expected value of error\\
\item $\sigma^2$ -- error variance\\
\item $Se = \frac{TP}{TP +FN}$ -- sensitivity\\
\item $PPV =  \frac{TP}{TP +FP}$ -- positive predictive value\\

\end{itemize}
Values of these metrics for the main network can be seen in table \ref{base_table}.
\begin{table*}[!ht]
\centering
    \caption{Quality metrics for the base network. Values are averaged across 20 networks.}
    \label{base_table}
   
    \begin{tabular}{|p{1.7cm}|p{1.7cm}|p{1.7cm}|p{1.7cm}|p{1.7cm}|p{1.7cm}|p{1.7cm}|}
    \hline
      Value & P begin & P end & QRS begin & QRS end & T begin & T end \\ 
    \hline  
       Se (\%) & 95.20 & 95.39 & 99.51 & 99.50 & 97.95 & 97.56 \\
       PPV(\%) & 82.66 & 82.59 & 98.17 & 97.96 & 94.81 & 94.96 \\
       $m\pm\sigma$(ms) & $2.7 \pm ${21.9} & $-7.4 \pm ${28.6} & $2.6 \pm $12.4 & $-1.7 \pm $14.1 & $8.4 \pm $28.2 & $-3.1 \pm $28.2\\ 
    \hline
    \end{tabular}
\end{table*}
The performance of the base network is comparable to the quality of direct methods \cite{2018arXiv180903393K} and may seem relatively good, but in the following sections we will demonstrate that even though these formal metrics' values appear to be promising, they are not trustworthy.
They do not account for the degree of representation of various pathologies within the dataset for which ECG segmentation is performed. For example, if the majority of samples belongs to healthy patients, and if the segmentation algorithm handles the standard healthy case right, but not the pathological one, then its quality assessment will directly depend on the number of pathological or artifact-containing samples in the data set. 

So it is important to investigate the behavior of the network on pathological cases rather than relying on formal metrics.
\section{Main network trends}
In this section we describe some qualities that the base deep network demonstrates while being trained throughout the training dataset.
\subsection{Noise stability}
All else being equal, the presence of high-frequency noise does not interfere with the network’s ability to produce a correct annotation, nor does it influence the smoothness of its output signal. 

\begin{figure}[ht!]
\centering
\includegraphics[scale=0.5]{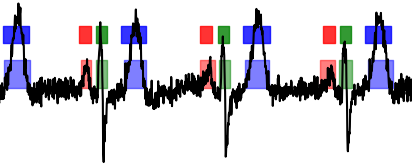}
\caption{Noised ECG is annotated correctly. The bottom row of colored boxes shows the annotation given by an expert, the top row depicts the annotation by the neural network, it's architecture is depicted in fig. \ref{convnet}}
\label{ecg_with_noise}
\end{figure}
Absence of necessity of noise filtration is an advantage of the deep learning approach, since it reduces the time spent on ECG preprocessing.

\subsection{Reaction to pathologies}
It turned out that the presence of pathology has the most noticeable effect on the quality of the neural network's performance.

\begin{figure}[ht!]
\centering
\includegraphics[scale=0.5]{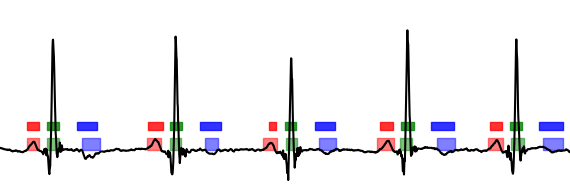}
\caption{This graph shows an abnormal case (containing unusual T-waves) being annotated by a base network with satisfactory results.}
\label{tbad}
\end{figure}

When analyzing the network's performance on a test sample of patients, it turned out that the following rule of thumb is valid:  if the case is pathological, it can still be properly annotated by a common neural network (for example, \ref{tbad} depicts an ECG with a non-standard T-wave shape and the network had annotated it well).
But if an ECG was erroneously annotated by the neural network, then this ECG is pathological (or unreadable due to artifacts of the recording process).

This is especially true for the QRS complex. The QRS complex turned out to be the easiest to mark up with a neural network (as well as direct algorithms). Typically, it has the largest signal amplitude, although this is not always true, for example see fig. \ref{ecg_with_noise}.

\begin{figure}[ht!]
    \vspace{1pt}
        \includegraphics[scale=0.35]{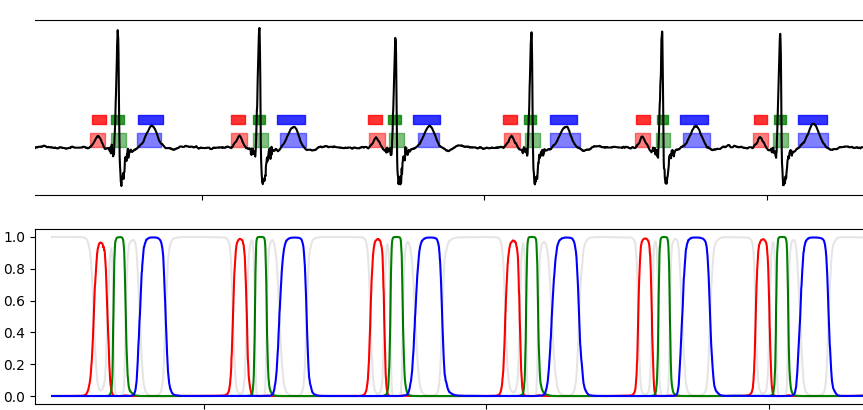}
    \caption{A simple case of ECG annotation. On the top graph, the bottom set of colored markings shows the doctor’s manual annotation for P waves (red), QRS complexes (green) and T waves (blue). The set of colored markings above represents the network’s annotation for the same ECG. The bottom graph shows the raw output signal of the network. These values represent the network’s “confidence” in the fact that the current segment does contain the appropriate ECG waveform. For a simple case of a healthy ECG, we can see that the network performs well when compared against a professional annotation done by a cardiologist. Smooth symmetric waves in the output signal of the network are characteristic of the segmentation of simple cases (like this one)}
\label{delin_example}
\end{figure}

\begin{figure}[ht!]
\centering
\includegraphics[scale=0.5]{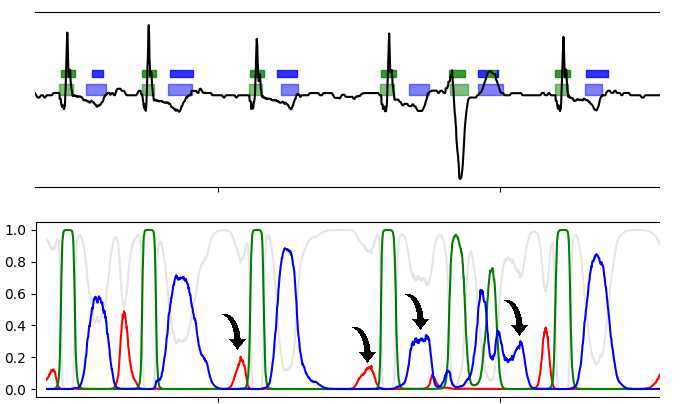}
\caption{ Distinctive features of the network’s output signal in a markedly pathological case. The top graph contains annotations for QRS complexes and T waves in green and blue respectively. The bottom set of markings represents the doctor’s annotation, the network’s annotation is located above. The P wave (red) is lacking completely, a fact which was both noted by the cardiologist and the network. Non-smooth asymmetric waves of arbitrary amplitude in the output signal of the network are characteristic of segmentation of an electrocardiogram with severe pathology (like in this case) and are shown by black arrows.
}
\label{worst}
\end{figure}

If we study the raw output of the network for an ECG which does not contain any noticeable pathologies and compare it against the output for a markedly pathological ECG, we can notice a systematic difference. In the pathological case, the network’s signal contains numerous asymmetric low-amplitude spikes, which do not contribute to the resulting annotation (see pic.\ref{worst}). However, we do not observe this kind of behavior in the non-pathological case. Should the channel contain a signal spike, it is smooth and has a large enough amplitude to influence the annotation (pic. \ref{delin_example}). In a sense, the intensity of this effect can be interpreted as the network's "confidence".

\section{Ensemble}
The ensemble formation procedure was designed in such a way that adding a new network to the ensemble fixes some errors of the already existing networks on the training set.

After training, the F-score of the base network was measured on each patient of the training sample, so that one could see in which cases the network does not perform well. Then, the procedure of iterative ensemble building starts. All patients rated at an F-score of 0.99 and above were removed from the training set. The rest of the training set is then used as a separate training set for the new neural network. 

This new neural network is then trained on that training set and, again, the procedure of screening out patients is carried out: all the cases on which this neural network has failed to achieve a score of 0.99 remain while others are deleted. The procedure described is repeated until the patients in the training set run out.

At each iteration of this algorithm, a new neural network is created. Each of these networks is trained on an ever decreasing data set. If, after one step, the sample size has not changed, then the same network is re-trained on the same sample, on the assumption that it fell into a bad local minima.

Figure \ref{stages} demonstrates an example of how the size of the training set has changed during the procedure described.
\begin{figure}[h!]
\centering
\includegraphics[scale=0.6]{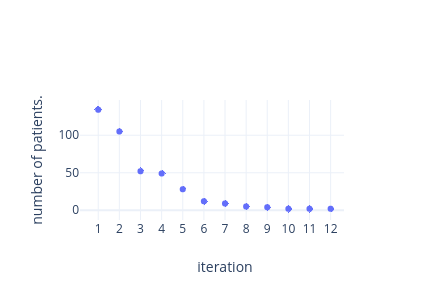}
\caption{Number of patients remaining in the train subset at every stage of ensemble formation. The formation continues until the number of patients in the subset reaches zero.}
\label{dynamics}
\end{figure}\\
When the ensemble is created, the resulting annotation for every input ECG can be obtained by averaging the raw output signals across all members of the ensemble.
\section{Error correction in action}
\begin{figure}[ht!]
\centering
\includegraphics[scale=0.5]{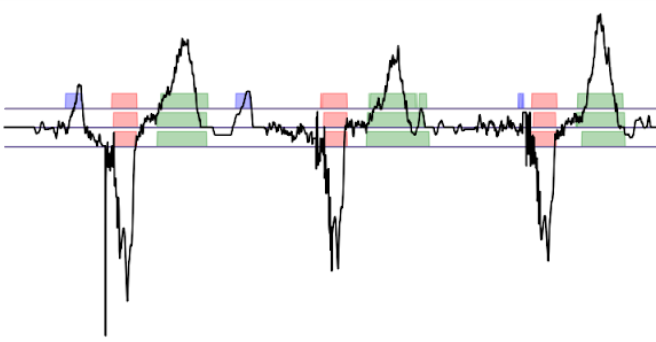}
\caption{Noticeable improvement of annotation quality through consecutive stages of ensemble training. The doctor’s annotation is shown at the bottom, the 4th network’s annotation is located in the middle and the 3rd network’s annotation is at the top. The 3rd network mistakenly notes the P-wave, blue. The remaining components of the cycle (green and red colors) are defined by both networks correctly.}
\label{stages}
\end{figure}

From the very ensemble construction algorithm itself, one can see that the ensemble is able to systematically correct some errors of the base network. Also, members of the ensemble can correct each other's mistakes. To illustrate that, we provide are a couple of typical examples: fig. \ref{stages} demonstrates how the 4th member of ensemble fixes the error of the 3rd network in case of an abnormal ECG.

\begin{figure}[h!]
\centering
\includegraphics[scale=0.50]{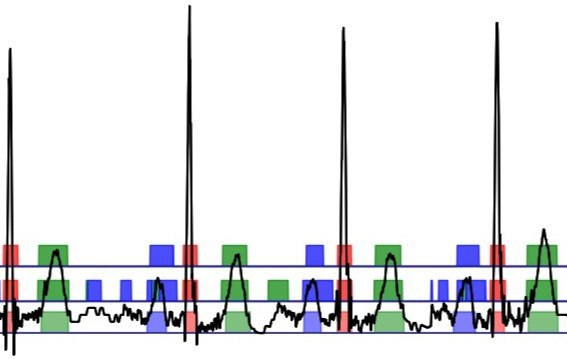}
\caption{This figure shows the annotation for an abnormal case. The doctor’s annotation is shown at the bottom, a single network’s annotation is located in the middle and the ensemble’s annotation is at the top. The improved quality of segmentation since the introduction of the ensemble is clearly visible.}
\label{ensemble_improves}
\end{figure}

Fig. \ref{ensemble_improves} depicts how the ensemble itself fixes an error of the base network for an abnormal case (patient was taken from test set).

Networks added to the ensemble at later iterations were trained on very small patient subsets (for example, on fig. \ref{dynamics} it is clearly visible that at least half of networks were trained on less than 10 patients). Every network has 60,804 trainable parameters. This situation obliges us to check for the degree of overfitting in such networks. In order to do so, we have designed a simple procedure, which roughly evaluates a degree of overfitting of all but one networks in an ensemble without need to use the test sample.
The key is that every member of the ensemble (except for the first one) is only trained on some subset of the training set of patients. This allows them to use the unseen part of the training set as their test set and therefore to evaluate their generalization ability.

\begin{figure}[h!]
\centering
\includegraphics[scale=0.50]{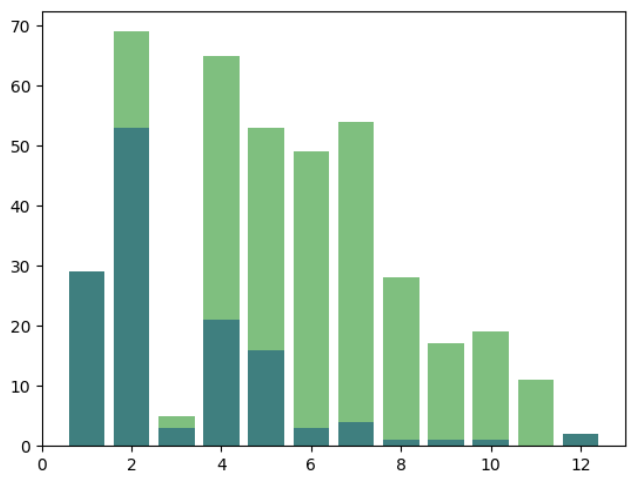}
\caption{This image demonstrates the generalization capability of every individual member of the ensemble. The ensemble members capability to annotate it’s part of the dataset exceptionally well( i.e. F-score not less then 0.99) is shown in dark green, while it’s ability to annotate the previously unseen part of the dataset exceptionally well is shown in light green.}
\label{speciality}
\end{figure}

Surprisingly, experimental results show that there are ensemble members that can produce good annotations (F-score higher than 0.99) for tens of unseen patients despite having been trained on 2-3 patients. Picture \ref{speciality} illustrates the behavior of an ensemble of 12 members: for every member of the ensemble, the dark green bar shows the amount of well annotated patients from the training set of that member, and the light green bar illustrates the amount of well annotated patients from the unseen part of the whole training set. We can clearly see that there are only two networks which are probably significantly overfitted (the first network doesn’t have its own test set, so we cannot make any judgments about it).
\\

\begin{figure}[!ht]
\centering
    \begin{subfigure}[t]{0.5\textwidth}
    \includegraphics[scale=0.46]{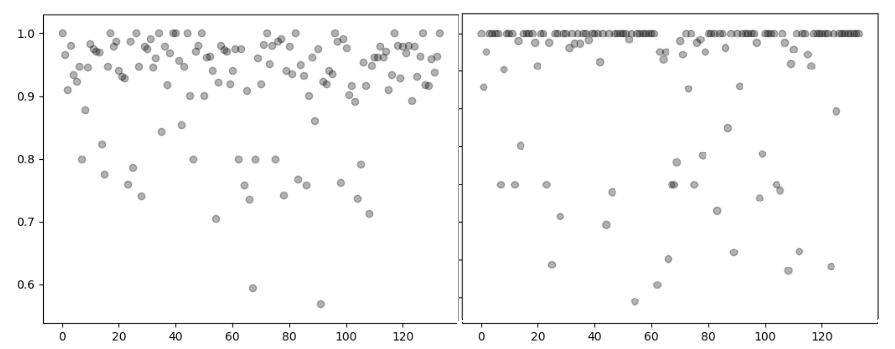}
    \label{train-ensemb}
    \end{subfigure}
    \\ 
    \begin{subfigure}[t]{0.5\textwidth}
    \includegraphics[scale=0.46]{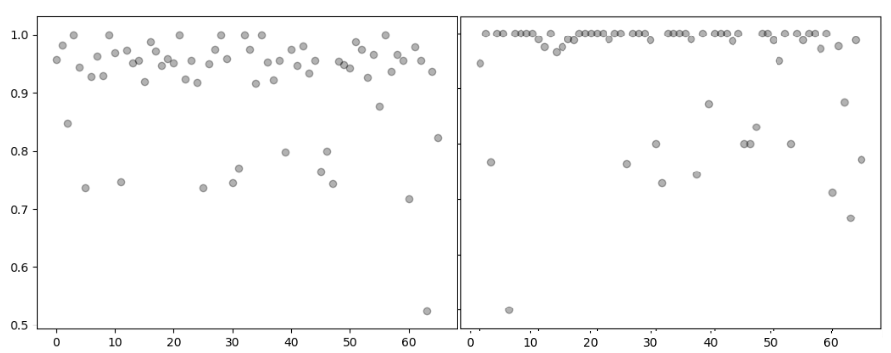}
    \label{test-ensemb}
    \end{subfigure}
\caption{F-score scattergrams for every patient demonstrates distillation effect. The horizontal axis corresponds to the patients, the vertical axis corresponds to the F-score of segmentation given by an ensemble (or) the base network. Top left: the base network the annotates training set. Top right: the ensemble annotates training set. Bottom left and bottom right:  the annotation for the test part when performed by the base network and an ensemble respectively}
\label{vis_behavior}
\end{figure}

To get an idea of the behavior of an ensemble built according to the principle described above, we visualized its behavior on a test and training set. We then compared this behavior to the behavior of the underlying network. The resulting visualization is shown in Fig. \ref{vis_behavior}. 

The left subfigures demonstrate how the base network annotates testing and training sets, while the right part depicts how the ensemble performs on them. Answers of the ensemble are more concentrated in a very narrow area of high F-scores near the 0.99 value. The shape of point clouds is the same for both the testing and the training sets, which means the behavior of the ensemble on the training set can be generalized. 

\begin{figure}[ht!]
\centering
\includegraphics[scale=0.60]{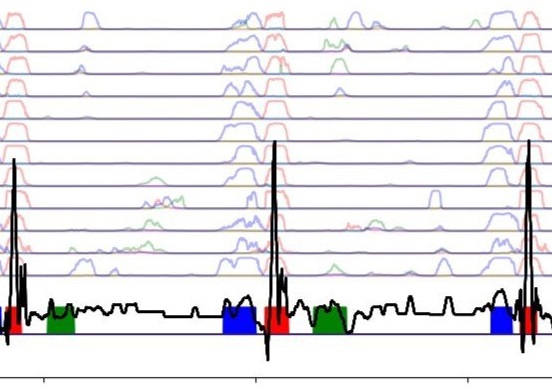}
\caption{An example of an outlier ECG on which both -- the ensemble and the base network -- fail: all the members have failed to segment the T-vawe (green), potentially because the amplitude of the T-wave has become comparable in scale to the amplitude of the noise. Colored blocks indicate the markup of the expert. The top markings show the output signals of the ensemble members. }
\label{outlier}
\end{figure}

Another conclusion is that outlier patients still exist even for the ensemble. An example of an outlier ECG  is depicted in pic. \ref{outlier}.  
\\

Moreover, outliers are now better highlighted than before. After the ensemble processing, the data set turned out to be divided into two classes: a superconcentrated cloud and a very rarefied one. Interestingly, despite the fact that the overall F-score of the ensemble is higher than that of the base network (for the base network it is 0.94 and for the ensemble is more then 0.95), the proportion of outliers detected by the ensemble is not less (and is even slightly higher) than that for the base network.

This "distillation" effect, in some cases, may possibly help to assess which types of errors can be corrected by searching for other local minima for this model, and which will most likely require other architectures or other training procedures.
\section{Conclusion}
The ensemble generation procedure allowed us to consider simultaneously many different local minima of the loss function. The differences between the minima were significant in the sense that the networks were required to give significantly different markup for subsets of ECGs. As expected, this has led to an improvement of the results of the base network. However, these improvements are not uniform across the dataset, which is interesting. A substantial group of samples is distinguished, which were annotated with a base network with minor errors (F-score higher then 0.9), and after using the ensemble, they became annotated close to the ideal (F-score close to 1). There is also a group of samples, which the base network annotated with errors (usually significant) and the ensemble does not improve a situation for them (moreover, it often worsens the segmentation of the base network in such a case). 

This leads to the assumption that the cases from the second set are fundamentally difficult for the considered neural network with adopted training procedure. Surprisingly, the presence of pathology on the ECG does not guarantee that this ECG will fall to the second set. It depends on the type of pathology, which means that among the pathologies there are "simple" and "complex" ones.

\section{Discussion}
The number of samples from the aforementioned class number two can be used as a basis for assessing the quality of data representation that a network of a given architecture can build. In the case considered, for example, it can be concluded that data representation built by the base network has serious flaws, despite the fact that formal quality metrics (such as specificity, positive predictive value and F-measure) show relatively high values.
Also, a promising area of research could be the search for a reliable metric that assesses the quality of network representation while doesn't consider the network as a black box. The creation of such metrics probably requires the further development of mathematical theory for deep learning. 
\bibliography{references}

\end{document}